# A semantic space for modeling children's semantic memory

Guy Denhière, CNRS (UMR 6146) & University of Aix-Marseille1, France

Benoît Lemaire, Leibniz-IMAG (CNRS UMR 5522), University of Grenoble, France

Cédrick Bellissens, University of Paris VIII, France

Sandra Jhean-Larose, IUFM Paris & University of Paris VIII, France

**A preliminary model based on a child corpus**

In contrast to many LSA semantic spaces in the literature which are based on domain-specific corpora, we chose to build a general child corpus intended to model children's semantic memory (Denhière, Lemaire, Bellissens & Jhean-Larose, 2004) and to offer a layer of basic semantic associations on top of which computational models of children cognitive processes can be designed and simulated (Lemaire, Denhière, Bellissens & Jhean-Larose, to appear).

*Developmental data*

We possess little direct information on the nature and the properties of the semantic memory of the children from 7 to 11 year-old (Howe & Courage, 2003; Murphy, McKone & Slee, 2003; Towse, Hitch & Hutton, 2002; De Marie & Ferron, 2003; Cycowicz, 2000). However we can refer to studies relating to the vocabulary acquisition for this period of cognitive development to build an approximation of children's semantic memory and to work out a corpus corresponding roughly to oral and written linguistic materials that a 7-11 year-old child is exposed to (Lambert & Chesnet, 2002; Lété, Springer-Charolles & Colé, 2004).

From a theoretical point of view, the question: "How children learn the meaning of words?" has received different kinds of answers. Bloom (2000) argued that the mind does not have a module for language acquisition, that the processes of association and imitation are not sufficient to explain word learning that requires rich mental capacities: conceptual, social, and linguistic, that interact in complicated ways. As Gillette, Gleitman, Gleitman & Lederer (1999) pointed out, "lexical and syntactic knowledge in the child, far from developing as separate components of an acquisition procedure, interact with each other and with the observed world in a complex, mutually supportive, series of bootstrapping operations" (p. 171). In connectionist modeling of language acquisition (see Elman, 2004; Shultz, 2003), Li, Farkas, & Mac Whinney (2004) developed DevLex, a self-organizing neural network model of the development of the lexicon, designed to combine the dynamic learning properties of connectionist networks with the scalability of represental models such HAL (Burgess & Lund, 1997) or LSA (Landauer & Dumais, 1997). Finally, as Louwerse & Ventura (2005) emphasized, children likely do not learn the meaning of words through learning words, but through different forms of discourse and by understanding their relation in context. In consequence, these authors assume that "LSA is how all language processing works and that LSA can be considered a model of how children learn the meaning of words" (p. 302).

From these points of view, whatever the complexity of the processes implicated in language acquisition, it should be possible to reproduce different states of meaning acquisition by estimating two things: the mean normative vocabulary of children and the written data children are exposed to. Then it should be possible to construct corpora trained in a model which take into account complex interaction of usage contexts.

One has estimated the quantitative aspects of vocabulary development. Landauer & Dumais (1997) wrote that "it has been estimated that the average fifth grade child spends about 15 minutes per day reading in school and another 15 out of school reading books, magazines, mail and comic books (Anderson, Wilson, & Fielding, 1988; Taylor, Frye & Maruyama, 1990). If we assume 30 minutes per day total for 150 school days and 15 minutes per day for the rest of the year, we get an average of 21 minutes per day. At an average reading speed of 165 words per minute, which may be an overestimate of natural, casual rates, and a nominal paragraph length of 70 words, they read about 2,5 paragraph per minute, and about 50 per day."

Ehrlich, Bramaud du Boucheron, and Florin (1978), estimated the mean vocabulary of 2.538 French children in 4 scholar grades (from $2^{nd}$ through $5^{th}$) and from 4 social and economic classes (senior executives, medium ranking executives, employees and workers) by using a judgement task of lexical knowledge (scaled in five levels, from "never heard" to "I know that word very well and I use it frequently") and a definition task ("tell all the senses of that word"). The verbal materials was made of 13.500 root words (63 % substantives, 17% verbs and 20% adjectives and adverbs) assumed to be representative to a general adult vocabulary (First year's college students).

Then, the 13.500 root words were judged by 150 adults on a five levels scale and the 2.700 better known and more frequently used words were taken to make part of the experimental materials submitted to children, to test the extent of their vocabulary as a function of age and social and economic classes. The results showed that the number of totally unknown root words decreased of about 4.000 from $2^{nd}$ through $5^{th}$ grade, the number of very well known root words increased of only 900 and the number of medium known root words increased of approximativelly 3.000. According to the authors, the frequently used vocabulary does not vary so much over primary school grades, but children are learning a lot of new words that, for a major part, they are unlikely to use (Ehrlich, Bramaud du Boucheron & Florin, 1978). The meaning knowledge level of substantives and verbs were equivalent whatever the age of the children, and were higher than knowledge level of adjectives and adverbs. Finally, the mean vocabulary was larger for favored classes (i.e., a mean difference of 600 words between the two extreme classes) and these differences kept the same over the 4 scholar grades. Moreover, after the results from the definition task, the vocabulary of the children was enriched by approximativelly 1.000 root words each school grade, the increase from $2^{nd}$ though $5^{th}$ grade was about 3.000 (Ehrlich, Bramaud du Boucheron & Florin, 1978).

More recently, Biemiller (2001, 2003), Biemiller & Boote (2005), Biemiller & Slonim (2001) presented results that confirmed most of the pionner results of Ehrlich et al. (1978). In short, Biemiller & Slonim (2001), referring to the Dale and O'Rourke's Living Word Vocabulary (1981), estimated that in the $2^{nd}$ grade the mean normative vocabulary was 5.200 root words, increasing to approximately 8.400 root words by 5<sup>th</sup> grade. This reflects acquiring about 2.2 words per day from ages 1 through 8 (end of grade two) and 2.4 words per day during ages 9 through 12. In other words, an average child learns between 800 and 900 root words a year, a figure that is close to the one obtained by Anglin (1993).

There is evidence that the vocabulary is acquired in largely the same order by most children. Biemiller and al. (2001) have found that when vocabulary data are ordered by children's vocabulary levels rather than their grade level, they can clearly identify ranges of words *known well* (above 75%), words *being acquired* (known between 25% and 74%) and words *little known*. At any given point in vocabulary acquisition, a child is likely to be learning root words from about 2.000 to 3.000 words in a sequence of 13.000 to 15.000 words. This makes the construction of a "vocabulary curriculum" plausible. Unfortunately, although these findings imply the existence of a well-defined sequence of word acquisition, a complete sequential listing of the 13,000–15,000 root words expected at the level of twelfth grade cannot now be furnished (for more details, see Biemiller, 2003, 2005).

*Our corpus*

Various kinds of written data children are exposed to have to be represented in our corpus. We gathered texts for a total of 3.2 million words. We could have gathered many more texts but we are concerned with cognitive plausibility. Our goal was to build a semantic space which reproduces as close as possible the verbal performances of a 7 to 11 year-old child. Moreover, we are not only concerned with mimicking the representation of the semantic memory, but also its development, from raw written data to the semantic representation of words. Therefore, the type and size of the input are important factors.

Children are exposed to various kinds of written data: storybooks, schoolbooks, encyclopedias, etc. It is very hard to estimate the proportion of each source to the total exposure to print. Actually, the main problem is that children have been exposed to language long before they can read: they learn the meaning of some words through exposure to speech, not to mention the perceptual input. This is a well-known limitation of LSA (Glenberg & Robertson, 2000) but it is less of a problem with adult corpora. Indeed, the proportion of words learned from reading is much higher for adults than for children. Therefore, we need to take this problem into account in the design of child corpora by trying to mimic this kind of input. We could have used spoken language intended for children, but these kinds of data are less formal than written language and thus harder to process. We ended up with gathering children's productions, a kind of language which is closer to basic spoken language than stories or textbooks. In addition, we decreased the significance of textbooks and the encyclopedia, because these sources of information affect the children's knowledge base after they have learned to read.

Altogether, the child corpus consists of stories and folk tales written for children (~1.6 million words), children's own written productions (~800,000 words), reading textbooks (~400,000 words) and a children's encyclopedia (~400,000 words).

The size of the corpus is another important factor. It is very hard to estimate how many words we process every day. It is not easy to estimate this number from the literature since, as we mentioned earlier, most studies are based either on just root words or on the number of words children know and not those they were exposed to. Then, according to Landauer & Dumais (1997) estimate, we consider a relevant corpus size to be tens of million words for adults, and several million words for children around 10 years of age. The size of the corpus we are presenting in this paper is 3.2 million words. After processing this corpus by LSA, we compared it with human data.

**Comparison of semantic similarities with human data**

In order to validate the specific child semantic space *Textenfants*, we used four more and more constraining tests: association norms, semantic judgments, a vocabulary test and memory tasks. First, we investigated whether our semantic space could account for word associations produced by children who were provided with words varying in familiarity. Therefore, the first test compares the performances of *Textenfants* on French verbal association norms recently published by de La Haye (2003). Second, we compared the performances of *Textenfants* with grades 2 and 5 children's judgments of semantic similarity between couples of words extracted from stories. Third, we tested whether our semantic space is able to represent definitions. LSA is often criticized for its ability to account for syntagmatic associations but not paradigmatic ones (French & Labiouse, 2002). Like Landauer et al. (1998) did with the TOEFL test, we compared LSA scores to performances of four groups of children. The task consisted in choosing the correct definition from a set of four definitions (correct, close, distant and unrelated). We believe that this test is more constraining than the first one because definitions could be either words, phrases or sentences but no dictionary was part of this corpus. Fourth, we used the semantic space to assess the children's performances of recall and summarization. We revisited results of nine prior experiments and compared the number of propositions recalled to the cosine measure between the source text and the text produced by each participant. To the extent that these are correlated *Textenfants* can be used to assess verbal protocols without resorting to the tedious propositional analysis.

*Associations norms*

The first experiment is based on verbal association norms published by de La Haye (2003). Because of previous results are based on children from grades 2 to 5, our interest here concerns the 9 year-old norms. Two-hundred stimulus printed words (144 nouns, 28 verbs and 28 adjectives) were provided to 100 9-years-old children. For each word, participants had to write down the first word that came to their mind. The result is a list of words, ranked by frequency. For instance, given the word *eau* (*water*), results are the following:

- boire (*drink*): 22%
- mer (*sea*): 8%
- piscine (*swimming pool*): 7%
  ...
- vin (*wine*): 1%
- froid (*cold*): 1%
- poisson (*fish*): 1%

This means that 22% of the 9-year old children provided the word *boire* (*drink*) when given the word *eau* (*water*). This value can be used as a measure of the strength of the association between the two words. These association values were compared with the LSA cosine between word vectors: we selected the three highest-ranked words as well as the three lowest-ranked (vin, froid, poisson in the previous example). We then measured the cosines between the stimulus and the highest-ranked, the 2nd highest, the 3rd-highest, and the mean cosine between the stimulus word and the three lowest-ranked. Results are presented in Table 1.

Table 1: Mean cosine between stimulus word and various associates for 9-years-old children

| Words | Mean cosine with stimulus word |
|---|---|
| Highest-ranked words | .26 |
| 2nd highest-ranked words | .23 |
| 3rd highest-ranked words | .19 |
| 3 lowest-ranked words | .11 |

*Student* t-tests show that all differences are significant ($p < .05$). This means that our semantic space is not only able to distinguish between the strong and weak associates, but can also discriminate the first-ranked from the second-ranked and the latter from the third-ranked.

The correlation with human data is also significant ($r(1184)=.39$, $p<.001$). Actually, two factors might have decreased this value. First, although we tried to mimic what a child has been exposed to, we could not control the frequencies with which each word occurred in the corpus. Therefore, some words might have occurred with a low frequency, leading to an inaccurate semantic representation. When the previous comparison was performed on the 20% of words with lower LSA weights (those words for which LSA has the most knowledge), the correlation was much higher ($r(234) =.57$, $p<.001$).

The second factor is the agreement among participants: when most children provide the same answer to a stimulus word, there is high agreement, which means that both words are very strongly associated. However, there are cases when there is almost no agreement: for instance, the first three answers to the word *bruit* (*noise*) are *crier* (*to shout*) (9%), *entendre* (*to hear*) (7%) and *silence* (*silence*) (6%). It is not surprising that the model corresponds better to the children's data in case of a high agreement, since this denotes a strong association that should be reflected in the corpus. In order to select answers whose agreement was higher, we measured their entropy. When we selected 20% of the items with the lowest entropy, the correlation increased: $r(234)=.48$, $p<.001$.

We also compared these results with the several adult semantic spaces, a literature corpus and four French newspaper corpora. Results are presented in Table 2. In spite of much larger sizes, all adult semantic spaces correlate worse than the children's semantic space with the data of the participants in the study. Statistical tests show that all differences between the children model and the other semantic spaces are significant ($p<.03$).

**Table 2: Correlations between participant child data and different kinds of semantic spaces**

| Semantic space | Size (in million words) | Correlation with children data |
|---|---|---|
| Children | 3.2 | .39 |
| Literature | 14.1 | .34 |
| Le Monde 1993 | 19.3 | .31 |
| Le Monde 1995 | 20.6 | .26 |
| Le Monde 1997 | 24.7 | .26 |
| Le Monde 1999 | 24.2 | .24 |

All these results show that the degree of association between words defined by the cosine measure within the semantic space seems to correspond quite well to children's judgments of association.

*Semantic judgments*

A second test consisted in comparing *TextEnfants* with two groups of forty five children's (Grade 3 vs grade 5) results in a semantic distance judgment task (Denhière & Bourguet, in preparation). We asked participants to judge on a five points scale the semantic similarity between two words extracted from a story used in the Diagnos™ tests (Baudet & Denhière, 1989). For each of the 7 stories selected, all the possible noun-noun couples were constructed ("giant-woman", "forest-woman", "house-woman" and so on) and had to be judged. For each couple, we obtained an average similarity value for each grade that we compared with LSA cosines for the two words of the couples.

Judgments were significatively correlated between the two groups of children (ranged from .80 to .95 as function of stories) and correlations between LSA cosines and children's judgements were all but one ("Giant" story) significant (see table 3).

**Table 3: Correlations between LSA cosines and children's semantic similarity judgments.**

|  | Grade 3 | Grade 5 | Couples (N) |
|---|---|---|---|
| Giant | 0,22 | 0,21 | 66 |
| Donkey | 0,24* | 0,22* | 105 |
| Truck | 0,28* | 0,25* | 66 |
| Chamois | 0,34* | 0,40** | 55 |
| Clowns | 0,48** | 0,50** | 56 |
| Lion | 0,55** | 0,63** | 66 |
| Bear cub | 0,55** | 0,58** | 91 |
| Mean | 0,37** | 0,37** |  |

(* p <.05; ** p <.01).

In conclusion, the degree of association between words (test 1) and the semantic distance between concepts (test 2), defined by the cosine measure within the semantic space *TextEnfants* fit quite well to children's productions and judgments.

*Vocabulary test*

The third experiment is based on a vocabulary test composed of 120 questions (Denhière, Thomas, Bourguet & Legros, 1999). Each question is composed of a word and four definitions: the correct one, a close definition, a distant definition and an unrelated definition. For instance, given the word *pente* (*slope*), translations of the four definitions are:

- *tilted surface which goes up or down (correct);*
- *rising road (close);*
- *vertical face of a rock or a mountain (distant);*
- *small piece of ground (unrelated).*

Participants were asked to select what they thought was the correct definition. This task was performed by four groups of 30 children from 2$^{nd}$, 3$^{rd}$, 4$^{th}$ and 5$^{th}$-grades. These data were compared with the cosines between a given word and each of the four definitions. Altogether, 116 questions were used because the semantic space did not contain four rarely occurring words.

Figure 1 displays the average percentage of correct, close, distant and unrelated answers for the 2$^{nd}$ and 5$^{th}$-grades. The first measure we used was the percentage of correct answers (LSA curve). It is .53 for the model, which is exactly the same value as for the 2$^{nd}$-grade children. Except for unrelated answers, the model data generally follow the same pattern as the children's data. When restricted to the 99 items that LSA had encountered frequently enough in the corpus (weight < .7), results of the model fall in-between the percent correct for 2$^{nd}$ and 3$^{rd}$-grade children.

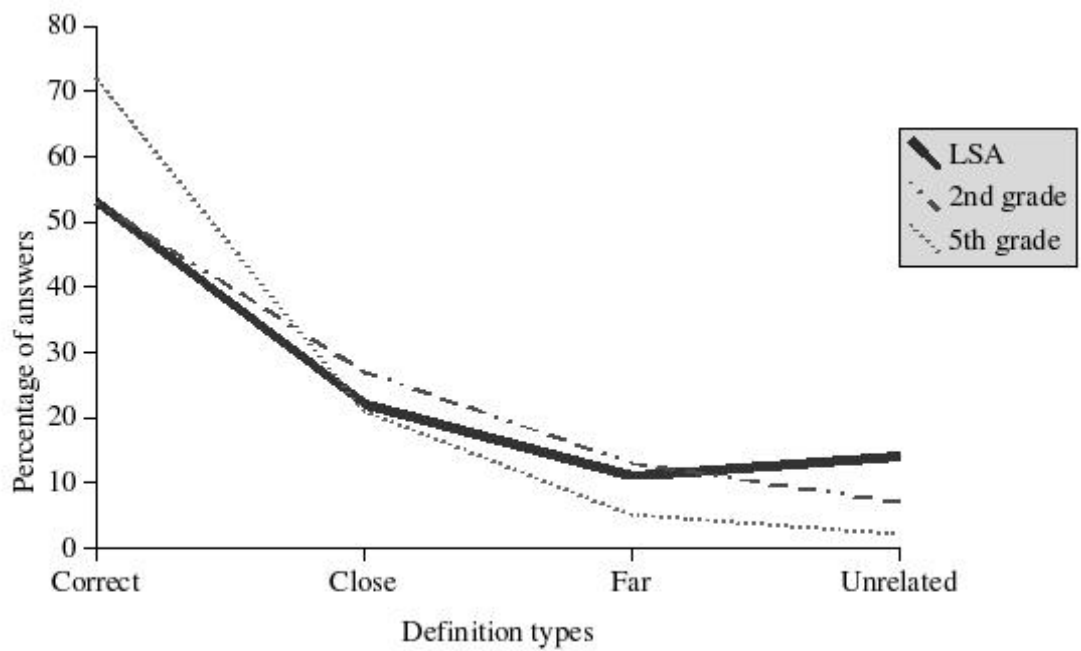

Figure 1. Percentage of answers for children and model data.

We also investigated a possible effect of corpus lemmatization. French uses more word forms than English. This fact could reduce the power of LSA because there might be less opportunity to encounter relevant contexts. Replacing all words by their lemma could therefore improve the representation of word meaning: instead of having $N_1$ paragraphs with word $P_1$, $N_2$ paragraphs with word $P_2$,... $N_p$ paragraphs with word $P_n$, we could have $N_1+N_2+...+N_p$ paragraphs with word P, which could be better from a statistical point of view. All forms of a verb are therefore represented by its infinitive, all forms of a noun by its nominative singular, and so forth. We used the Brill tagger on French files developed by the CNRS-ATILF laboratory at Nancy, France, as well as the Namer lemmatizer. However, results were worse than before when all words were lemmatized. Verb lemmatization also resulted in no improvment . Hence, lemmatization proved to be of no help in the present case.

Then, we modified the lemmatizer in order to lemmatize only verbs and found better results, almost exactly corresponding to the 2$^{nd}$-grade data. This can be due to the fact that verb lemmatization groups together forms that have the same meaning. However, the different forms of a noun do not have exactly the same meaning. For instance, *flower* and *flowers* are not arguments of the same predicates. In our semantic space, the cosine between *rose* and *flower* is .51 whereas the cosine between *rose* and *flowers* is only .14. *Flower* and *flowers* do not occur in the exact same contexts. Another example can be given from the neighbors of *bike* and *bikes*. First three neighbors of *bike* are *handlebar*, *brakes* and *pedals* whereas first three neighbors of *bikes* are *motorbikes*, *trucks* and *cars*. *Bike* and *bikes* do not occur either in the same contexts. Therefore, representing them by the same form probably confuses LSA.

*Memory tasks*

The last experiment is based on recall or summary tasks (Thomas, 1999). Three groups of children (mean age of 8,3 years) and six groups of adolescents (between 16 and 18 years old) were asked to read a text and write out as much as they could recall, immediately after reading or after a fixed delay of one week. For three of these groups, participants were asked to write a summary. We used seven texts. We tested the ability of the semantic representations to estimate the amount of knowledge recalled. This amount is classically estimated by means of a propositional analysis: first, the text as well as each participant production are coded as propositions. Then, the number of text propositions that occur in the production is calculated. This measure is a good estimate of the information recalled from the text. Using our semantic memory model, this amount of recall is given by the cosine between the vector representing the text and the vector representing the participant production.

Table 4 displays all correlations between these two measures. They are all significant (< .05) and range from .45 to .92, which means that the LSA cosine applied to our children's semantic space provides a good estimate of the text information recalled, whatever the memory task, recall or summary, and the delay.

**Table 4: Correlations between LSA cosines and number of propositions recalled for different texts.**

| Texts | Task | Number of participants | Correlations |
|---|---|---|---|
| *Hen* | Immediate recall | 52 | .45 |
| *Dragon* | Delayed recall | 44 | .55 |
| *Dragon* | Summary | 56 | .71 |
| *Spider* | Immediate recall | 41 | .65 |
| *Clown* | Immediate recall | 56 | .67 |
| *Clown* | Summary | 24 | .92 |
| *Bear cub* | Immediate recall | 44 | .62 |
| *Bull* | Delayed recall | 23 | .69 |
| *Giant* | Summary | 105 | .58 |

In an experiment with adults, Foltz (1996) has shown that LSA measures can be used to predict comprehension. Besides validating our model of semantic memory, this experiment shows that an appropriate semantic space can be used to assess text comprehension more quickly than propositional analysis, which is a very tedious task.

**Corpus stratification for developmental studies**

The corpus presented so far contains the kind of texts a child is exposed to, but it does not reproduce the order in which texts are processed over a child's life. Actually, it would be very valuable to have a corpus in which paragraphs are sorted according to the age they are intended for, which would make it possible to obtain a corpus for virtually any age. Comparing the model of semantic memory (or any processing model based on it) with data from children would be much more precise. In addition, it would then be possible to simulate the development of semantic memory.

We proceeded in two steps: first, we gathered more texts that were intended for a specific age and second, we sorted every paragraph of each category according to a readability measure.

*Gathering texts for specific age*

We defined four age levels: 4 to 7 years, 7 to 11 years, 11 to 18 years and adults (over 18 years). The goal was to build a well-rounded semantic space that includes lexical, encyclopedic and factual information. Lexical information is given by dictionaries. Encyclopedic information is provided by schoolbooks and encyclopedia. Factual information is obtained from stories and novels. For each age level, we therefore collected French texts in each of the following categories: children's written productions, dictionary, encyclopedia, textbooks and stories or novels. Each level is included in the next one. Level 1 is therefore composed of productions from 4 to 7 year-old children, a dictionary and an encyclopedia for 4 to 7 year-old children, 1st grade reading textbooks and stories for 4 to 7 year-old children. Level 2 is composed of all the texts of level 1 plus productions from 7 to 11 year-old children, a dictionary and an encyclopedia for 7 to 11 year-old children, reading texts and stories for 7 to 11 year-old children, etc. For adults, the daily newspaper *Le Monde* was used in place of textbooks.

*Sorting paragraphs according to a readability measure*

In order to be more precise, all paragraphs of a given level were sorted according to the age they best correspond to. We relied on a readability measure defined by Mesnager (1989, 2002), which has been carefully standardized. This measure is based on the percentage of difficult words and the mean length of sentences. Difficult words are those that are not part of a list of 32,000 French common words. The formula is the following:

*readability = 3/4 percentage of difficult words + 1/4 mean sentence length*

This measure is very rough, but we used it merely as a tool for sorting all paragraphs within each level. Thus we computed the readability measure for every paragraph of each category. This work is very preliminary. As a rough test, we found that for each information source, the mean measure for level N was higher than the mean measure for level N-1, which is satisfactory.

**Applications of the model of semantic memory**

In this section, we present two applications of this model of semantic memory. The first one consists in comparing the relationships between co-occurrence frequency and LSA similarities. The second one aims at linking this model of semantic memory to a model of comprehension based on the construction-integration model designed by Kintsch (1988, 1998).

*Studying the development of semantic similarities*

The semantic similarity of two words (or, stated differently, their associative strength) is classically reduced to their frequency of co-occurrence in language: the more frequently two words appear together, the higher their similarity. This shortcut is used as a quick way of estimating word similarity, for example, in order to control the material of an experiment, but it also has an explanatory purpose: people tend to judge two words as similar because they were exposed to them simultaneously. Undoubtedly, the frequency of co-occurrence is correlated with human judgments of similarity (Spence & Owens, 1990). However, several researchers have questioned this simple relation (Bellissens & Denhière, 2002; Burgess, Livesay & Lund, 1998). The goal of this simulation is to use our semantic space to study the relation between co-occurrence and similarity.

An ideal method to study the relation between co-occurrence and similarity would consist in collecting all of the texts subjects have been exposed to and comparing their judgments of similarity with the co-occurrence parameters of these texts, a task that is obviously impossible. One could think of a more controlled experiment, by asking participants to complete similarity tests before and after text exposure. The problem is that the mental construction of similarities through reading is a long-term cognitive process, which would probably not be apparent over a short period.

It is also possible to count co-occurrences in representative corpora, but that would give only a global indication a posteriori. Thus, we would learn nothing about the direct effect of a given first or second-order co-occurrence on semantic similarity. It is useful to know precisely the effect of direct and high-order co-occurrences during word acquisition. Assume a person X who has been exposed to a huge set of texts since learning to read. Let S be the judgment of similarity of X between words W1 and W2. The questions we are interested in are:

- what is the effect on S of X reading a passage containing W1 but not W2?

- what is the effect on S of X reading a passage containing W1 and W2?

- what is the effect on S of X reading a passage containing neither W1 nor W2, but words co-occurring with W1 and W2 (second-order co-occurrence)?

- what is the effect on S of X reading a passage containing neither W1 nor W2, but third-order co-occurring words?

Our simulation follows the evolution of the semantic similarities of 28 pairs of words over a large number of paragraphs, according to the occurrence values. We started with a corpus size of 2,000 paragraphs. We added one paragraph, ran LSA on this 2001-paragraph corpus and, for each pair, computed the gain (positive or negative) of semantic similarity due to the new paragraph and checked whether there were occurrences, direct co-occurrences or high-order co-occurrences of the two words in the new paragraph. We then added another paragraph, ran LSA on the 2002-paragraph corpus, etc. Each new paragraph was simply the following one in the original corpus. More precisely, for each pair X-Y, we put each new paragraph into one of the following categories:

- occurrence of X but not Y;
- occurrence of Y but not X;
- direct co-occurrence of X and Y;
- second-order co-occurrence of X and Y, defined as the presence of at least three words which co-occur at least once with both X and Y in the current corpus;
- third- or-more-order co-occurrence, which forms the remainder (no occurrence of X or Y, no direct co-occurrence, no second-order co-occurrence). This category represents three- or-more co-occurrences because paragraphs whose words are completely neutral with X and Y (that is they are not linked to them by a high-order co-occurrence relation) do not modify the X-Y semantic similarity.

We stopped the computation at the 13,637th paragraph. 11,637 paragraphs were thus traced. This experiment took three weeks of computation on a 2 GHz computer with 1.5 Gb RAM. As an example, Figure 2 describes the evolution of similarity for the two words *acheter* (*buy*) and *magasin* (*shop*). This similarity is -.07 at paragraph 2,000 and increases to .51 at paragraph 13,637. The curve is quite irregular: there are some sudden increases and decreases. Our next goal was to identify the reasons for these variations.

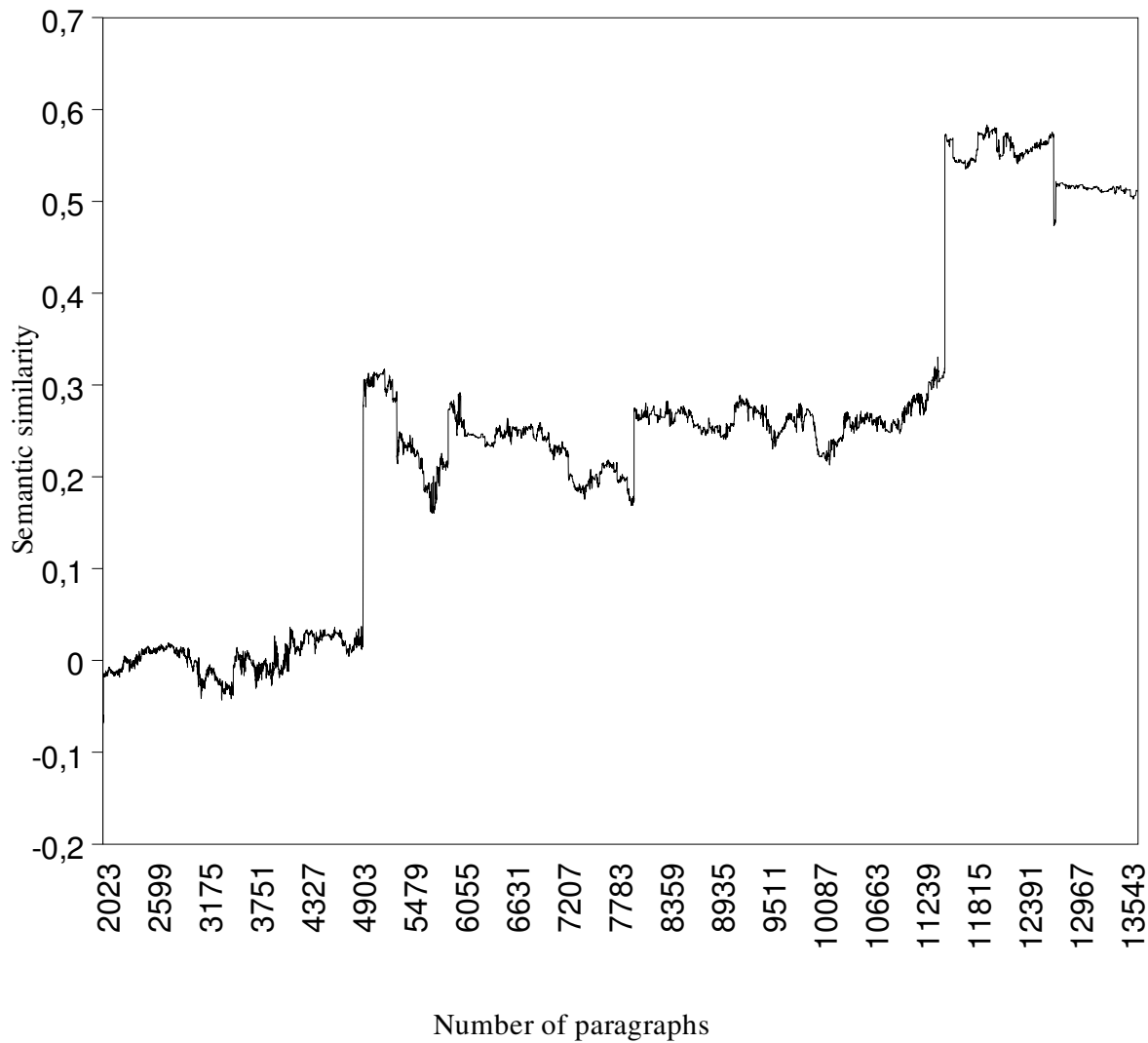

Figure 2. Similarity between *acheter (buy)* and *magasin(shop)* according to the number of paragraphs.

For each pair of words, we partialed out the gains of similarity among the different categories. For instance, if the similarity between X and Y was .134 for the 5,000-paragraph corpus and .157 for the 5,001-paragraph corpus, we attributed the .023 gain in similarity to one of the five previous categories. We then summed up all gains for each category. Since the sum of the 11,637 gains in similarity is exactly the difference between the last similarity and the first one, we ended up with a distribution of the total gain in similarity among all categories. For instance, for the pair *acheter(buy)-magasin(shop)*, the .58 (.51 - (-.07)) total gain in similarity is partialed out in the following way:

- -.10 due to occurrences of only *acheter(buy)*;
- -.19 due to occurrences of only *magasin(shop)*;
- .73 due to the co-occurrences;
- .11 due to second-order co-occurrences;
- .03 due to third-or-more-order co-occurrences.

This means that the paragraphs containing only *acheter(buy)* contributed all together to a decrease in similarity of .10. This is probably due to the fact that these occurrences occur in a context which is different to the *magasin(shop)* context. In the same way, occurrences of *magasin(shop)* led to a decrease in the overall similarity. Co-occurrences tend to increase the similarity, which is expected, and high-order co-occurrences contribute also to an increase.

We performed the same measurement for all 28 pairs of words. These pairs were selected from the 200 items of the association task presented earlier and their first-ranked associated word, as provided by children. We kept only words that appeared at least once in the first 2,000 paragraphs in order to have the same number of semantic similarities for all pairs. Average results are the following:

- -.15 due to occurrences of the first word;
- -.19 due to occurrences of the second word;
- .34 due to the co-occurrences;

- .05 due to second-order co-occurrences;
- .09 due to third-or-more-order co-occurrences.

First of all, we found pairs of words that never co-occur (e.g. *farine(flour)-gâteau(cake)*) even though their semantic similarity increases. Another result is that, except in a few cases, the gain in similarity due to a co-occurrence is higher than the total gain in similarity. This result occurs because of a decrease due to occurrences of only one of the two words (-.15 and .-19). In addition, high-order co-occurrences play a small but significant role: they tend to increase the similarity (.14 in total).

*Modeling text comprehension on the basis of the semantic memory model*

In a second application, we use our semantic memory model as a component of a fully automatic model of text comprehension, based on the construction-integration (CI) model (Kintsch, 1998), a predication mechanism (Kintsch, 2001, Lemaire & Bianco, 2003) and a model of the episodic buffer. This model is implemented in a Perl program that takes as input a text divided into propositions (Lemaire et al, to appear). Figure 3 shows the general architecture.

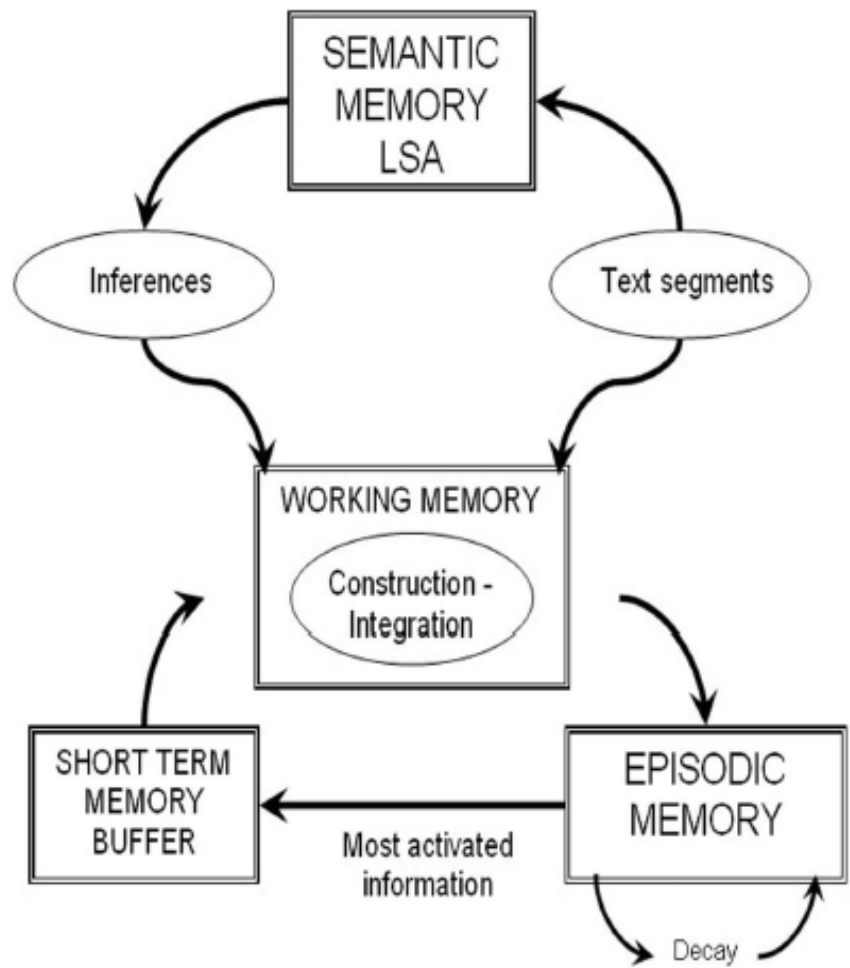

Figure 3. Architecture of the model of text comprehension.

Let us explain the flow of information which is an operational approximation of the CI model. Each proposition is considered in turn. First, the LSA semantic space (also called semantic memory) is used to provide terms that are semantically similar to the propositions and their components, thus simulating the process by which we automatically activate word associates in memory while reading. A fixed number of associates is provided for the predicate, but also for each of the arguments. For instance, given the proposition *carry(truck, food)* corresponding to the sentence *The truck is carrying bikes*, our semantic space provides the terms *transport*, *kilometers* and *travel* for the proposition, the terms *car*, *garage* and *vehicle* for the argument *truck* and the terms *motorbikes*, *trucks* and *cars* for the argument *bikes*. Associates for the arguments are simply the terms whose cosine with the arguments in the semantic space is higher. Associates for the predicate require a bit more calculation since only those that are close enough (beyond a given threshold) to at least one of the arguments are kept. This algorithm comes from Kintsch's predication model (Kintsch, 2001). In our example, the three associates *transport*, *kilometers* and *travel* were kept because they are similar enough to the *truck* or *bikes*.

Not all associates are relevant within the context, but this is not a problem because the next phase will rule out those that are not related to the current topic. Actually, all the terms in working memory (none for the first sentence) are added to the current proposition and all its associates. As we will explain later, relevant propositions coming from episodic memory can also be added. All these terms and propositions are processed by the construction/integration phase of Kintsch's model. The LSA semantic space is used once more to compute the similarity weights of links between all pairs of terms or propositions. For instance, the weight of *truck/bikes* is .67, the weight of *carry(truck, bikes)/travel* is .16, etc. Next, a specific spreading activation mechanism is used to give activation values to terms according to the weight of their connection with other terms. Those that are the most strongly linked to other terms will receive high activation values. For instance, the term *garage* in the previous example is weakly linked to all other terms or propositions. Therefore, this term will be given a low association value and will be dropped out.

The most highly activated items are stored in working memory. Three strategies are available:

- either a fixed number of items is selected, following Baddeley's model (2000); or
- a maximum sum of activation is defined, which is distributed among the items (Just & Carpenter, 1992); or
- only those items whose activation values are above a given threshold are selected.

Selected elements will be added later to the next proposition and its associates. They are also stored in episodic memory. This memory is the list of all terms or propositions already encountered, either because they were part of text propositions, or because they were provided by semantic memory as associates. Terms or propositions are attached to an activation value in episodic memory. This value is initially their activation value in working memory, but a decay function lowers this value over time. This activation value is increased each time a new occurrence of a term or proposition is stored in episodic memory. Both values are then combined according to functions we will not explain here. What is important is that elements in episodic memory can be recalled by the construction phase in case they are similar enough to the current proposition. The semantic space is then used once more to compute similarities and to decide whether elements in episodic memory are recalled or not. This last process simulates the fact that there was a digression in the text, leading to the fact that the previous propositions were dropped out from working memory. When the digression ends, these propositions will be recalled from episodic memory because they will be similar again to the current propositions. However, if the digression lasts too long, the decay function will have greatly lowered the activation values of the first propositions and they will not be recalled any more.

Here is an English translation of a simulation we performed with the children semantic space *Textenfants*. Consider the following text: *The gardener is growing his roses. Suddenly, a cat meows. The man throws a flower at it*.

The first proposition is *grow(gardener,roses)*. Terms that are neighbors of *grow* but also close to *gardener* or *roses* are: *vegetable, vegetables, radish*. Neighbors of *gardener* are: *garden*, *border*, *kitchen garden* (one word in French). Neighbors of *roses* are: *flowers*, *bouquet*, *violets*. LSA similarities between all pairs of words are computed. After the integration step, working memory is:

– *grow(gardener,roses)* (1.00)

– *grow* (.850)

– *gardener* (.841)

– *border* (.767)

– *garden* (.743)

– *roses* (.740)

– *flowers* (.726)

Terms that are not related to the context (like *radish*) were removed.

The second proposition is *meow(cat)*. Since the second sentence is not related to the first one, no terms appear to be gathered from episodic memory. Terms that are neighbors of *meow* but also close to *cat* are: *meows*, , *purr*. Neighbors of *cat* are *meow*, *meows*, *miaow*. They are added to working memory. LSA similarities between all pairs of words are computed. After the integration step, working memory is:

- *meow(cat)* (1.00)
- *grow(gardener,roses)* (.967)
- *cat* (.795)
- *meow* (.776)

All terms related to the first sentence (*border*, *garden*, ...) were removed from working memory, because the second sentence is not related to the first one. However, the entire proposition is still there.

The third proposition is *throw(man,flower)*. *Flowers*, *bouquet*, *roses* and *violets* are back to working memory, because they are close to the argument *flower*. Terms that are neighbors of *throw* but also close to *man* or *flower* are: *command*, *send*, *Jack*. *Man* is too frequent for providing good neighbors; therefore the program did not consider it. Neighbors of *flower* are: *petals*, *pollen*, *tulip*. LSA similarities between all pairs of words are computed. After the integration step, working memory is:

- *flower* (1.00)
- *flowers* (.979)
- *petals* (.978)
- *grow(gardener,roses)* (.975)
- *roses* (.974)
- *violets* (.932)
- *bouquet* (.929)
- *throw(man,flower)* (.917)

- *tulip* (.843)

- *pollen* (.832)

To sum up, the LSA semantic space is used three times: first, to provide associates to the current proposition and its arguments; second, to compute the weights of links in the construction phase and third, to compute similarities between the current proposition and the episodic memory items for their possible reactivation in working memory.

The main parameters of that model are:

- the minimal and maximal weight thresholds for terms, in order to only consider terms for which LSA has enough knowledge;
- the number of associates provided by the LSA semantic space;
- the similarity threshold for the predication algorithm;
- the strategy for selecting items in working memory after the integration step;
- the decay and updating functions of episodic memory.

This program can be used with any semantic space, but, when combined with the semantic space described in this paper, it should be very useful for the study of children text comprehension.

**Conclusion**

In this paper, we presented a semantic space that is designed to represent children's semantic memory. It is based on a multi-source corpus composed of stories and tales, children's productions, reading textbooks and a children's encyclopedia. Our goal is to build a general semantic space as opposed to domain-specific ones. Results of the four tests we performed (comparison with association norms, semantic judgments, vocabulary test, recall and summary tasks) are promising (Denhière & Lemaire, 2004). The next step is to simulate the development of this semantic memory in order to reproduce developmental changes in performance on these tests among children aged 4 to 7 years to adolescence. One of the goals would be to reproduce the effect of prior knowledge on text comprehension (Caillies, Denhière & Jhean-Larose, 1999; Caillies, Denhière & Kintsch, 2002).

The semantic space we constructed was linked to a comprehension model derived from the construction-integration model (Kintsch, 1998). Like the CI model, our model uses semantic memory to activate the closest neighbors of the current concept or proposition, to select the most relevant elements and to keep the most activated in working memory. In addition, several models of short-term memory are implemented and an episodic buffer is used to store ongoing information and provide relevant prior items to working memory. Our current work consists in adding a long-term working memory (Ericsson & Kintsch, 1995) and a generalization process based on memory traces from the episodic buffer (Bellissens & Denhière, 2004; Denhière, Lemaire, Bellissens & Jhean-Larose, 2004).

**Acknowledgements**

We would like to thank Eileen and Walter Kintsch and an anonymous reviewer for their comments on a previous version of this paper.